\pdfoutput=1

\documentclass[11pt]{article}
\usepackage[]{ACL2023}

\usepackage{times}
\usepackage{latexsym}

\usepackage[T1]{fontenc}

\usepackage[utf8]{inputenc}

\usepackage{microtype}

\usepackage{inconsolata}

\usepackage{booktabs}

\usepackage{microtype}
\usepackage[utf8]{inputenc}
\usepackage[T1]{fontenc}
\usepackage{times}
\usepackage{latexsym}
\usepackage{amsfonts}
\usepackage{amsmath}
\usepackage{amssymb}
\usepackage{amsthm}
\usepackage{xspace}
\usepackage{comment}
\usepackage{array}
\newcolumntype{?}{!{\vrule width 2pt}}
\usepackage{inconsolata} 

\usepackage[group-separator={,},group-minimum-digits={3}]{siunitx}

\usepackage{cleveref}
\crefname{section}{\S}{\S\S}
\Crefname{section}{\S}{\S\S}
\crefname{table}{Tab.}{}
\crefname{figure}{Fig.}{}
\crefname{algorithm}{Algorithm}{}
\crefname{equation}{eq.}{}
\crefname{appendix}{App.}{}
\crefname{thm}{Theorem}{}
\crefname{prop}{Proposition}{}
\crefname{cor}{Corollary}{}
\crefname{observation}{Observation}{}
\crefname{assumption}{Assumption}{}
\crefformat{section}{\S#2#1#3}

\usepackage{tikz}
\usetikzlibrary{bayesnet}
\usetikzlibrary{arrows}
\usepackage{color}
\usepackage{graphicx}
\usepackage{caption}
\usepackage{subcaption}
\usetikzlibrary{backgrounds}

\usepackage{hhline}
\usepackage{xcolor}
\usepackage{colortbl}
\definecolor{Grey}{RGB}{210,210,210}

\usepackage{bbm}

\newcommand{\setquote}[1]{``{#1}''}

\def\expect{{\mathbb{E}}}

\def\vb{{\mathbf{b}}}

\def\vdelta{{\boldsymbol{\delta}\xspace}}

\def\vpi{{\boldsymbol{\pi}\xspace}}

\def\vomega{{\boldsymbol{\omega}\xspace}}
\def\vkappa{{\boldsymbol{\kappa}\xspace}}
\def\vlambda{{\boldsymbol{\lambda}\xspace}}

\newcommand{\normal}{\mathrm{Normal}}
\newcommand{\gam}{\mathrm{Gamma}}
\newcommand{\zerogeometric}{\mathrm{Zero \text{-} infl. \text{ } Geom.}}

\newcommand{\betadist}{\mathrm{Beta}}
\newcommand{\categorical}{\mathrm{Categorical}}
\newcommand{\ordered}{\mathrm{Ord}}
\newcommand{\ordnormal}{\mathrm{OrderedNormal}_{C}}
\newcommand{\dirichlet}{\mathrm{Dirichlet}}

\usepackage{tipa}
\newcommand{\ucambridge}{\normalfont \text{\textipa{D}}}
\newcommand{\ethz}{\text{\normalfont \textipa{Q}}}
\newcommand{\mitinst}{\normalfont \text{\textipa{@}}}
\newcommand{\epfl}{\normalfont \text{\textipa{N}}}
\newcommand{\uchicago}{\normalfont \text{\textipa{R}}}

\usepackage{tikz}

\definecolor{Sdark}{RGB}{12, 18, 194}
\definecolor{Pdark}{RGB}{133, 12, 194}
\definecolor{Qdark}{RGB}{194, 12, 133}
\definecolor{Odark}{RGB}{12, 179, 194}

\newcommand{\Sdark}[1]{{\color{Sdark} #1}}
\newcommand{\Qdark}[1]{{\color{Qdark} #1}}
\newcommand{\Pdark}[1]{{\color{Pdark} #1}}
\newcommand{\Odark}[1]{{\color{Odark} #1}}

\newcommand*\doubleCircle{\tikz[baseline=(char.base)]{
    \node[shape=circle,draw,inner sep=2pt,scale=0.8,double distance=0.35mm] (char) {\color{white}1};}}

\newcommand{\OrdNorm}{Ordered Normal\xspace}

\newcommand{\sN}{n}
\newcommand{\bN}{N}
\newcommand{\sC}{c}
\newcommand{\sZ}{z_{n}}

\newcommand{\sZbar}{\bar{z}_{n}}
\newcommand{\sZdot}{\dot{z}_{n}}

\newcommand{\sS}{\Sdark{s_{n}}}
\newcommand{\sQ}{\Qdark{q_{n}}}
\newcommand{\sP}{\Pdark{p_{n}}}
\newcommand{\sO}{\Odark{o_{n}}}

\newcommand{\bC}{C}
\newcommand{\bS}{S}

\newcommand{\bO}{O}

\newcommand{\bZvec}{\bar{\boldsymbol{z}}_{\ell}}
\newcommand{\bQvec}{\Qdark{\bar{\boldsymbol{q}}_{\ell}}}
\newcommand{\bPvec}{\Pdark{\bar{\boldsymbol{p}}_{\ell}}}

\newcommand{\TTZ}{\boldsymbol{\alpha}^{(z)}}
\newcommand{\TZ}{\vpi^{(z)}}
\newcommand{\TSz}{\Sdark{\vpi^{(s)}_{z_{n}}}}
\newcommand{\TSc}{\Sdark{\vpi^{(s)}_{c}}}
\newcommand{\TS}{\Sdark{\boldsymbol{\Pi^{(s)}}}}

\newcommand{\TPomegaz}{\Pdark{\omega^{(p)}_{z_{n}}}}
\newcommand{\TPkappaz}{\Pdark{\kappa^{(p)}_{z_{n}}}}
\newcommand{\TPomegaOne}{\Pdark{\omega^{(p)}_{1}}}
\newcommand{\TPomegac}{\Pdark{\omega^{(p)}_{\sC}}}
\newcommand{\TPomegaC}{\Pdark{\omega^{(p)}_{\bC}}}
\newcommand{\TPomegacMinusOne}{\Pdark{\omega^{(p)}_{\sC-1}}}
\newcommand{\TPkappac}{\Pdark{\kappa^{(p)}_{\sC}}}
\newcommand{\TPomega}{\Pdark{\vomega^{(p)}}}
\newcommand{\TPkappa}{\Pdark{\vkappa^{(p)}}}

\newcommand{\TQdeltaz}{\Qdark{\delta^{(q)}_{z_{n}}}}
\newcommand{\TQbz}{\Qdark{b^{(q)}_{z_{n}}}}
\newcommand{\TQdeltac}{\Qdark{\delta^{(q)}_{\sC}}}
\newcommand{\TQbc}{\Qdark{b^{(q)}_{\sC}}}
\newcommand{\TQdelta}{\Qdark{\vdelta^{(q)}}}
\newcommand{\TQb}{\Qdark{\vb^{(q)}}}

\newcommand{\TOz}{\Odark{\vpi^{(o)}_{z_{n}}}}
\newcommand{\TO}{\Odark{\boldsymbol{\Pi^{(o)}}}}

\newcommand{\sppd}{\text{SPPD}}
\newcommand{\lppd}{\text{LPPD}}

\newcommand{\bX}{X}

\newcommand{\bY}{Y}

\newcommand{\sY}{y_{n}}
\newcommand{\sYhat}{\widehat{y}_{n}}
\newcommand{\sYremain}{y_{n}^{\text{remain}}}

\usepackage[normalem]{ulem}

\title{An Ordinal Latent Variable Model of Conflict Intensity}

\author{
Niklas Stoehr$^{\ethz}$~\;~Lucas Torroba Hennigen$^{\mitinst}$~\;~\;Josef Valvoda$^{\ucambridge}$\\ ~\;~\textbf{Robert West}$^{\epfl}$~\;~\textbf{Ryan Cotterell}$^{\ethz}$~\;~\textbf{Aaron Schein}$^{\uchicago}$
\\
$^{\ethz}$ETH Z{\"u}rich \quad $^{\mitinst}$MIT \quad $^{\ucambridge}$University of Cambridge \quad $^{\epfl}$EPFL \quad $^{\uchicago}$The University of Chicago\\
\footnotesize \href{mailto:niklas.stoehr@inf.ethz.ch}{\texttt{niklas.stoehr@inf.ethz.ch}} \qquad \href{mailto:lucastor@mit.edu}{\texttt{lucastor@mit.edu}} 
\qquad \href{mailto:jv406@cam.ac.uk}{\texttt{jv406@cam.ac.uk}}\\
\footnotesize \href{mailto:robert.west@epfl.ch}{\texttt{robert.west@epfl.ch}} \qquad \href{mailto:ryan.cotterell@inf.ethz.ch}{\texttt{ryan.cotterell@inf.ethz.ch}}
\qquad \href{mailto:schein@uchicago.edu}{\texttt{schein@uchicago.edu}}
}

\begin{document}
\maketitle
\begin{abstract}
Measuring the intensity of events is crucial for monitoring and tracking armed conflict. Advances in automated event extraction have yielded massive data sets of \setquote{who did what to whom} micro-records that enable data-driven approaches to monitoring conflict. The Goldstein scale is a widely-used expert-based measure that scores events on a conflictual--cooperative scale. It is based only on the action category (\setquote{what}) and disregards the subject (\setquote{who}) and object (\setquote{to whom}) of an event, as well as contextual information, like associated casualty count, that should contribute to the perception of an event's \setquote{intensity}. To address these shortcomings, we take a latent variable-based approach to measuring conflict intensity. We introduce a probabilistic generative model that assumes each observed event is associated with a latent intensity class. A novel aspect of this model is that it imposes an ordering on the classes, such that higher-valued classes denote higher levels of intensity. The ordinal nature of the latent variable is induced from naturally ordered aspects of the data (e.g., casualty counts) where higher values naturally indicate higher intensity. We evaluate the proposed model both intrinsically and extrinsically, showing that it obtains good held-out predictive performance.
\end{abstract}


\vspace{0.5em}
\hspace{0em}\includegraphics[width=1.25em,height=1.25em]{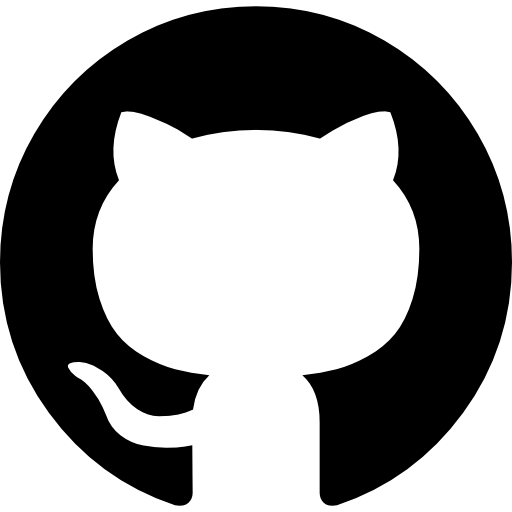}\hspace{.75em}\parbox{\dimexpr\linewidth-7\fboxsep-2\fboxrule}{\url{https://github.com/niklasstoehr/ordinal-conflict-intensity}}
\vspace{-0.5em}

\section{Introduction}

On a scale from $-10$ for conflictual to $+10$ for cooperative, which of the following events should be considered more \setquote{intense}: \setquote{Soldiers injured two civilians} or \setquote{Rebels detained fifty soldiers}?

\begin{table}[t]
\fontsize{10}{10}\selectfont
\centering
\renewcommand{\arraystretch}{1.0} 
\setlength{\tabcolsep}{0.1em} 
\begin{tabular}{@{\hskip 0.2cm}clcc@{\hskip 0.2cm}}
\toprule
\textbf{\begin{tabular}[c]{@{}c@{}}CAMEO\\ code\end{tabular}} & \multicolumn{1}{c}{\textbf{\begin{tabular}[c]{@{}c@{}}action\\ name\end{tabular}}} & \textbf{\begin{tabular}[c]{@{}c@{}}Goldstein\\ value\end{tabular}} & \textbf{\begin{tabular}[c]{@{}c@{}}avg. \#\\ casualties\end{tabular}} \\
\midrule
19                                                            & fight                                                                              & -10.0                                                              & 9.31                                                                  \\
20                                                            & mass violence                                                                      & -10.0                                                              & 42.20                                                                 \\
18                                                            & assault                                                                            & -9.0                                                               & 11.47                                                                 \\
15                                                            & force posture                                                                      & -7.2                                                               & 0.13                                                                  \\
17                                                            & coerce                                                                             & -7.0                                                               & 1.44                                                                  \\
14                                                            & protest                                                                            & -6.5                                                               & 2.06                                                                  \\
13                                                            & threaten                                                                           & -6.0                                                               & 0.13                                                                  \\
10                                                            & demand                                                                             & -5.0                                                               & 0.01                                                                  \\
12                                                            & reject                                                                             & -4.0                                                               & 0.00                                                                  \\
16                                                            & reduce relations                                                                   & -4.0                                                               & 0.00                                                                  \\
9                                                             & investigate                                                                        & -2.0                                                               & 0.04                                                                  \\
11                                                            & disapprove                                                                         & -2.0                                                               & 0.03                                                                  \\
1                                                             & public statement                                                                   & 0.0                                                                & 0.19                                                                  \\
4                                                             & consult                                                                            & 1.0                                                                & 0.03                                                                  \\
2                                                             & appeal                                                                             & 3.0                                                                & 0.00                                                                  \\
5                                                             & diplom cooperation                                                                 & 3.5                                                                & 0.03                                                                  \\
3                                                             & intent cooperate                                                                   & 4.0                                                                & 0.05                                                                  \\
8                                                             & yield                                                                              & 5.0                                                                & 0.10                                                                  \\
6                                                             & material cooperation                                                               & 6.0                                                                & 0.01                                                                  \\
7                                                             & provide aid                                                                        & 7.0                                                                & 0.00                                                                 \\
\bottomrule
\end{tabular}
\caption{The \href{https://parusanalytics.com/eventdata/cameo.dir/CAMEO.SCALE.txt}{The Goldstein scale} is an expert-based intensity ranking of the \num{20} \href{http://data.gdeltproject.org/documentation/CAMEO.Manual.1.1b3.pdf}{CAMEO action categories} ranging between $-10.0$ (conflictual) to $+10.0$ (cooperative). The scale disregards casualty counts that are typically considered in conflict assessment. Here, we display casualty counts as reported in the \href{https://dataverse.harvard.edu/dataset.xhtml?persistentId=doi:10.7910/DVN/INNYEO}{NAVCO dataset}.}
\label{tab:cameo_goldstein}
\end{table}

Measuring the intensity of events is crucial for monitoring and tracking armed conflict. Advances in the automated collection and coding of events have produced massive and systematized data sets of micro-records that enable data-driven approaches to monitoring conflict. While the ``intensity'' of a given event has traditionally been assessed by human expert raters, the tremendous quantity of events collected every day makes case-by-case analysis unmanageable. As a consequence, there is a strong demand for automated and model-based methods to aggregate events into meaningful ``conflict intensity'' measures.

One of the most frequently used measures is the Goldstein scale \citep{goldstein_conflict-cooperation_1992}. Major event datasets like IDEA~\citep{bond_integrated_2003}, KEDS~\citep{schrodt_kansas_2008}, GDELT~\citep{leetaru_gdelt_2013}, ICEWS~\citep{boschee_icews_2015}, Phoenix~\citep{beieler_creating_2016} and NAVCO~\citep{lewis_nonviolent_2016} all rely on it. The Goldstein scale assigns intensity scores between $-10.0$ and $+10$ on a conflictual--cooperative scale to the action categories defined by the \href{http://data.gdeltproject.org/documentation/CAMEO.Manual.1.1b3.pdf}{Conflict and Mediation Event Observations (CAMEO)} event coding scheme~\citep{schrodt_cameo_2012}. CAMEO specifies \num{204} low-level event types which are summarized into \num{20} high-level action categories. The Goldstein scale ranks \setquote{use unconventional mass violence} and \setquote{fight} as the most conflictual of the 20 high-level action categories ($-10.0$) and \setquote{provide aid} ($+7.0$) as the most cooperative; see \cref{tab:cameo_goldstein}.

Despite its usage, the Goldstein scale has many well-known shortcomings~\citep{king_automated_2003, schrodt_stuff_2019}. In particular, it applies only to action categories, and does not account for any contextual information of a given event, like which actors are involved, or how many fatalities resulted, among other bits of context that should contribute to the perception of an event's ``intensity''.~\looseness=-1

This paper takes a latent-variable based approach to measuring conflict intensity. We introduce a probabilistic generative model that assumes each observed event $n$ is associated with a latent intensity class $z_n$. A novel aspect of this model is that it imposes an ordering on the classes, such that higher values of $z_n$ denote higher levels of intensity. The ordinal nature of $z_n$ is induced from naturally ordered aspects of the data (e.g., casualty counts) where higher values naturally indicate higher intensity. The model effectively learns to interpolate the ordered (i.e., cardinal or ordinal) elements of the data while inferring correlation structure with the non-ordered (e.g., categorical) elements of the data (e.g., actor types).\looseness-1

We start with a discussion of the Goldstein scale and introduce a political event dataset annotated with Goldstein values in \cref{sec:problems} and \cref{sec:data}. Then, we propose our model with an ordinal latent variable in \cref{sec:model}. We evaluate the performance of the model intrinsically (\cref{sec:intrinsic_evaluation}) and extrinsically (\cref{sec:extrinsic_evaluation}) and find that it improves over measures based on the original Goldstein scale or heuristics based on the raw data.

\section{Limitations of the Goldstein Scale}
\label{sec:problems}

The Goldstein scale is a widely-used measure of the conflictual versus cooperative nature of interactions between countries~\citep{goldstein_conflict-cooperation_1992}. The scale was created by a panel of international relations experts who ranked descriptions of interactions. It was initially created to score action categories in the WEIS event ontology~\citep{mcclelland_world_1984} and was later adapted to CAMEO~\citep{schrodt_cameo_2012}.

The Goldstein scale applies only to the action category of an event (e.g., ``fight'' or ``trade''). Thus, two different ``fight'' events, which might involve two different pairs of actors, occur at different times, or differ dramatically with respect to the number of associated fatalities, will still be assigned the same Goldstein value. The Goldstein scale is thus a  poor measure of a conflict's perceived ``intensity'', as it ignores much of the information that contributes to that perception. Recent work in conflict studies, for instance, operationalizes ``intensity'' primarily using casualty counts \citep{chaudoin_beyond_2017,zhong_extracting_2023}, which the Goldstein scale ignores entirely.

In~\cref{tab:cameo_goldstein}, we show the empirical distribution of assigned Goldstein values alongside the empirical distribution of casualty counts in a dataset of conflict events. The Goldstein scale is very coarse-grained; while it ostensibly ranges between $-10.0$ and $+10.0$, only a small number of discrete values ever occur, with many action categories assigned the same value. For the purpose of measuring conflict intensity, a finer-grained and more contextual scale is desirable.

\section{Conflict Event Data}
\label{sec:data}

This paper considers the publicly available \href{https://dataverse.harvard.edu/dataset.xhtml?persistentId=doi:10.7910/DVN/INNYEO}{Nonviolent and Violent Campaigns and Outcomes (NAVCO)} data collection~\citep{chenoweth_days_2018}, specifically, the latest release NAVCO 3.0 from November 2019 which comprises $\bN = \num{112089}$ events between December 1990 and December 2012. An exemplary event description is \setquote{On 19 May 2012, \Sdark{soldiers} \Pdark{injured} \Qdark{two} \Odark{civilians} in Afghanistan}. Each part of this description has been parsed by human coders into standardized, structural features. We color-code the features that correspond to the semantic roles \Sdark{subject}, \Pdark{predicate}, \Qdark{quantifier}, \Odark{object}, which are the focus of our modeling approach. Each data point $n$ thus consists of a four-element tuple $\{ \sS, \sP, \sQ, \sO \}$. We note that events are further coded for their location (in this case, Afghanistan) and time (19 May 2012), among other bits of contextual information. Let us discuss each feature in more detail:

\begin{figure}[t]
     \centering
     \includegraphics[width=0.95\linewidth]{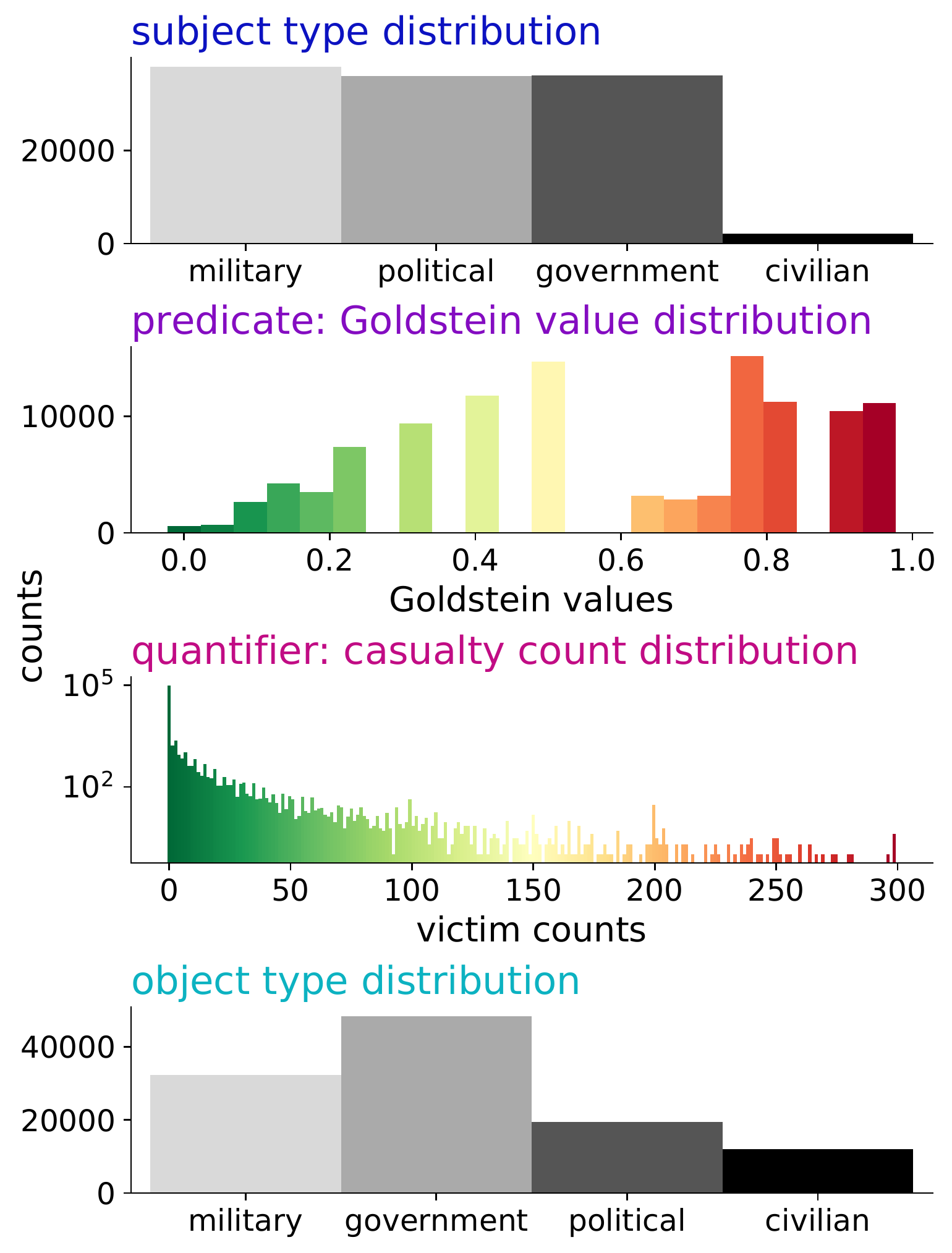} 
     \caption{Data distributions of \href{https://dataverse.harvard.edu/dataset.xhtml?persistentId=doi:10.7910/DVN/INNYEO}{NAVCO 3.0 dataset} \citep{chenoweth_days_2018}. Goldstein values and casualty counts have intrinsic intensity orderings. Goldstein values are reversed and transformed, so that \num{1.0} represents the most conflictual. We model the \Sdark{subject} $\sS$ as Categorical, the \Pdark{predicate} $\sP$ as Beta, the \Qdark{quantifier} $\sQ$ as Zero-inflated Geometric and the \Odark{object} $\sO$ as Categorical.~\looseness=-1}
     \label{fig:data_distribution}
 \end{figure}

\paragraph{\Sdark{Subject} $\sS$.}
 
NAVCO contains columns termed \setquote{actor3}, \setquote{actor6} and \setquote{actor9} which code for the subject (or agent) of a given action. The actor types are defined by the \href{http://data.gdeltproject.org/documentation/CAMEO.Manual.1.1b3.pdf}{CAMEO actor codebook}. We first merge the higher-level categories \setquote{actor3} and \setquote{actor6}, resulting in \num{33} different actor types, and then map all actor types into one of $S=4$ classes: $\sS \in \{\text{civilian}, \text{military}, \text{governmental}, \text{political}\}$. We present our \num{4}-class actor type mapping in~\cref{tab:actor_mappings} of the appendix. 
 
\paragraph{\Pdark{Predicate} $\sP$ (Goldstein values).}
 
NAVCO codes\footnote{NAVCO features a $21^{\textrm{st}}$ action category which we exclude since it is not specified by the \href{ http://data.gdeltproject.org/documentation/CAMEO.Manual.1.1b3.pdf
}{CAMEO taxonomy} and thus has no Goldstein value.} each event description into one of the \href{https://parusanalytics.com/eventdata/cameo.dir/CAMEO.SCALE.txt}{20 CAMEO action categories} in the column \setquote{verb10}, which is by extension associated with a Goldstein value $\sP$. Throughout, we refer to the Goldstein value $\sP$ as an action's ``predicate'', since there is a one-to-one mapping between action categories and Goldstein values. We scale Goldstein values $\sP$ to a $[0,1]$ range and invert them (i.e., $\sP \leftarrow 1-\sP$) so that higher values close to \num{1} represent more conflictual action categories.

\paragraph{\Qdark{Quantifier} $\sQ$ (casualty counts).} Each event description is annotated with human-verified fatality and wounded counts. We add the two and refer to the resulting value $\sQ \in \mathbb{N}_{0}^{+}$ as an event's ``quantifier'' or its ``casualty count''. In \cref{tab:cameo_goldstein}, we give the average number of casualties associated with each action alongside its Goldstein value---as intuition might suggest, actions that Goldstein scores as more conflictual (e.g., \setquote{fight} (\num{-10.0})) coincide with more casualties on average.~\looseness=-1

\paragraph{\Odark{Object} $\sO$.} Similar to its subject, NAVCO codes for the direct object or ``target'' of a conflict action using the CAMEO coding scheme; these codes are found in the columns \setquote{target3} and \setquote{target6}. We map those into the $\bO =$ \num{4} classes so that $\sO \in \{\text{civilian}, \text{military}, \text{governmental}, \text{political}\}$.
 
\paragraph{Contextual information: location and time.}

Finally, each event is further annotated with a timestamp and location, which we use to design extrinsic evaluation tasks in \cref{sec:extrinsic_evaluation}.~\looseness=-1

\section{Ordinal Latent Variable Model}
\label{sec:model}

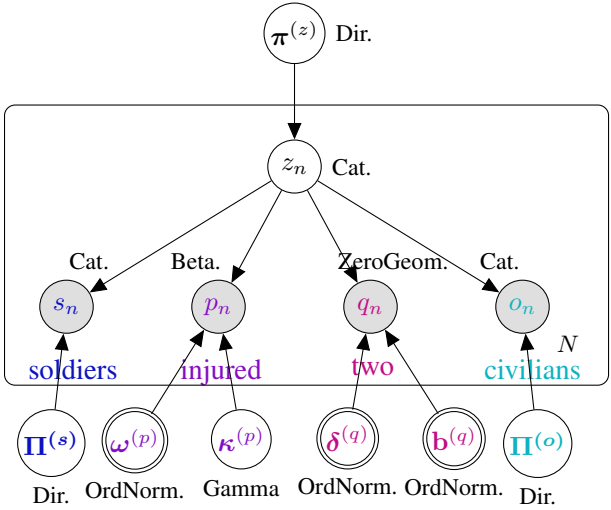
\begin{figure}[t]
\centering
\begin{tikzpicture}
 \node[obs,xshift=-3.0cm, text= Sdark, label={[shift={(0.3,0.0)}]Cat.}] (S) {$\sS$};%
 \node[obs,xshift=-1.0cm, text= Pdark, label={[shift={(-0.3,0.0)}]Beta.}] (P) {$\sP$};%
 \node[obs,xshift=1.0cm, text= Qdark, label={[shift={(0.3,0.0)}]ZeroGeom.}] (Q) {$\sQ$};%
 \node[obs,xshift=3.0cm, text= Odark, label={[shift={(-0.3,0.0)}]Cat.}] (O) {$\sO$};%
 \node[latent, xshift = 0cm, above = 1.5cm, label=right:{Cat.}] (Z) {$\sZ$};%

\node[text width=1cm, below=of S, yshift=0.8cm]
    {\Sdark{soldiers}};
\node[text width=1cm, below=of P, yshift=0.8cm]
    {\Pdark{injured}};
\node[text width=0.5cm, below=of Q, yshift=0.8cm]
    {\Qdark{two}};
\node[text width=1cm, below=of O, yshift=0.8cm]
    {\Odark{civilians}};
 
 \node[latent,above=of Z, label=right:{Dir.}] (TZ) {$\TZ$}; %
 \node[latent,below=of S, label=below:{Dir.}, xshift=-0.2cm] (TS) {$\TS$};
 \node[latent,below=of P, xshift=-1.1cm, double distance=0.35mm, label=below:{OrdNorm.}] (TPomega) {$\TPomega$}; %
 \node[latent,below=of P, xshift=0.3cm, label=below:{Gamma}] (TPkappa) {$\TPkappa$}; %
 \node[latent, below=of Q, xshift=-0.3cm, double distance=0.35mm, label=below:{OrdNorm.}] (TQdelta) {$\TQdelta$}; %
 \node[latent,below=of Q, xshift=1.1cm, double distance=0.35mm, label=below:{OrdNorm.}] (TQb) {$\TQb$}; %
 \node[latent,below=of O, label=below:{Dir.}, xshift=0.2cm] (TO) {$\TO$}; 

 \tikzset{plate caption/.style={caption, node distance=0, inner sep=-2pt,
below right=5pt and 100pt of #1.south,text height=0.2em,text depth=0.0em}}
\plate [inner sep=.45cm, xshift=0.0cm, yshift=0.0cm] {N} {(P)(Q)(O)(S)(Z)} {$N$}; %

 \edge {Z} {P, Q, O, S};
 \edge {TZ} {Z};
 \edge {TPomega} {P}; 
 \edge {TPkappa} {P}; 
 \edge {TQdelta} {Q};
 \edge {TQb} {Q};
 \edge {TO} {O};
 \edge {TS} {S};
\end{tikzpicture}

\caption{The proposed latent variable model of conflict intensity. 
The observed sites of an event tuple $\{ \sS, \sP, \sQ, \sO \}$ describe the \Sdark{subject type}, \Pdark{Goldstein value}, \Qdark{casualty counts}, \Odark{object type} in an event description such as \setquote{\Sdark{soldiers} \Pdark{injured} \Qdark{two} \Odark{civilians}}. The parameters $\TPomega$, $\TQdelta$ and $\TQb$ are ordinally transformed vectors, as indicated by double-border nodes (\doubleCircle) which leads the latent $\sZ$ to represent ordinal intensity classes.}
\label{fig:graphical_model}
\end{figure}

We operationalize conflict intensity as a latent variable that expresses the association between the observed variables \Sdark{subject} ($\sS$), \Pdark{predicate} ($\sP$), \Qdark{quantifier} ($\sQ$) and \Odark{object} ($\sO$). Each data point is a tuple $\{ \sS, \sP, \sQ, \sO \}$ representing an event. Our Bayesian latent variable model is depicted in \cref{fig:graphical_model}.
We assume the following generative story. For each event $\sN$, we assume that its event intensity class $\sZ \in \{1, \ldots, C\}$ is a Categorical random variable
\begin{align}
    \sZ &\sim \categorical(\TZ)
\end{align}
where $\TZ$ is a $C$-dimensional discrete distribution over latent intensity classes. We place a Dirichlet prior over $\TZ$
\begin{align}
\TZ &\sim \dirichlet(\TTZ)
\end{align}
with concentration parameter $\TTZ \in \mathbb{R}_{+}^{\bC}$. Conditioned on $\sZ$, we assume each of the observed sites per event tuple $\sS, \sP, \sQ$ and $\sO$ are then drawn as
\begin{align}
    \sS \mid \sZ &\sim \categorical(\TSz) \\
    \sP \mid \sZ &\sim \betadist(\TPomegaz, \TPkappaz) \label{eq:goldstein_p}\\
    \sQ \mid \sZ &\sim \zerogeometric(\TQdeltaz, \TQbz) \label{eq:quantifier_q}\\
    \sO \mid \sZ &\sim \categorical(\TOz)
\end{align}
Here $\TSz$ and $\TOz$ are the discrete distributions for class $z_n$ over $S$ subject and $O$ object types, respectively. They are given as row vectors in the matrices $\TS \in (0,1)^{\bC \times \bS}$ and $\TO \in (0,1)^{\bC \times \bO}$ that $\sZ$ indexes into. We place a simple symmetric Dirichlet prior over all rows---e.g., $\TSc \sim \dirichlet(\mathbf{1}_S)$ for $c \in \{1,\dots, C\}$. The scalar parameters $\TPomegaz, \TPkappaz, \TQdeltaz, \TQbz$ are similarly selected by $\sZ$ from $\bC$-dimensional vectors $\TPomega, \TPkappa, \TQdelta, \TQb$ as discussed below.

\subsection{Ordinal Latent Variable}
We want the latent variable $\sZ$ to be ordinal, such that higher-valued classes correspond to higher intensity levels. However, in the model thus described, $\sZ$ is a Categorical random variable whose classes are not inherently ordered. So how does this model encode an ordinal $z_n$? Unlike the subject $\sS$ and object $\sO$, which are categorical, the Goldstein value $\sP$ and casualty count $\sQ$ are cardinal quantities whose magnitudes naturally indicate the ``intensity'' of a given event. To capture this intuition, we first assume that $\sP$ and $\sQ$ are drawn from Beta (\cref{eq:goldstein_p}) and Zero-inflated Geometric distributions (\cref{eq:quantifier_q}), respectively, whose parameters are indexed by $\sZ$. We then impose an ordering on the parameters, such that higher classes (e.g., $\sZ=c$) correspond to higher-valued parameters (e.g., $\TPomegac > \TPomegacMinusOne$) which in turn encourage larger values of the observed cardinal quantities (e.g., $\sP$).~\looseness=-1\footnote{We note that one might also impose ordering on casualty types (e.g., an event with civilian casualties might be considered more \setquote{intense} than one with military casualties). In this work, however, we focus just on learning scales from observations that are naturally cardinal.}

\paragraph{\OrdNorm}
To flexibly impose ordering on vectors of parameters, we first define the \OrdNorm prior~\citep{stoehr_sentiment_2023}. An ordered vector $\vlambda = (\lambda_1, \ldots, \lambda_{\bC})$, where $\lambda_c > \lambda_{c-1}$, is a $C$-dimensional \OrdNorm random variable $\vlambda \sim \ordnormal(\mu, \sigma)$ if it is sampled according to the following generative process:
\begin{align}
    &x_{\sC} \stackrel{\textrm{iid}}{\sim}  \normal(\mu, \sigma) \quad \text{ for } \sC \in  \{1,\,\dots\,, \bC\} \nonumber\\ 
    &(\lambda_1,\dots,\lambda_{\bC}) \leftarrow \ordered(\{x_1,\dots,x_{\bC}\})
\end{align}
where $\ordered(\cdot)$ takes in an unordered set of numbers $\{x_1,\dots,x_{\bC}\}$, and transforms it into a vector whose components are in strictly increasing order,
\begin{align}
\lambda_c \leftarrow
\begin{cases}
    x_1 & \text{if } c = 1\\
    x_1 + \sum_{i = 2}^c \exp({x_i}) & \text{if } c > 1
\end{cases}
\label{eq:orderedTransform}
\end{align}

This transformation is invertible and differentiable and thus does not obstruct gradient-based inference of model parameters (\cref{app:inv-ordered-normal}).

\paragraph{Ordered Beta.}
We model Goldstein values $\sP$ as Beta random variables. For intensity class $\sZ = \sC$, we assume $\sP \sim \betadist(\TPomegac, \TPkappac)$, where $\TPomegac \in (0,1)$ is the mode and $\TPkappac > 0$ is the concentration parameter. We impose an ordering on the modes by positing a transformed OrderedNormal prior:
\begin{align}
S^{-1}(\TPomega) &\sim \ordnormal(\mu, \sigma)
\end{align}
where $\TPomega = (\TPomegaOne, \ldots, \TPomegaC)$ is the $\bC$-dimensional ordered vector of modes and $S^{-1}(\cdot)$ is the element-wise inverse sigmoid function. To ensure that each class-conditional Beta distribution is unimodal around its mode $\TPomegac$, we then mandate that the concentration parameter $\TPkappac$ is greater than 2 by imposing a shifted Gamma prior.
\begin{align}
(\TPkappac - 2)  &\stackrel{\textrm{iid.}}{\sim} \gam(k, \eta)
\end{align}
where $k$ and $\eta$ are the shape and rate parameters. We note that while the elements of $\TPomega$ are ordered, those of $\TPkappa$ are not.

\paragraph{Ordered Zero-inflated Geometric.}
We model the casualty counts $\sQ$ as Zero-inflated Geometric random variables. For intensity class $\sZ = \sC$, we assume $\sQ \sim \zerogeometric(\TQdeltac, \TQbc)$ where $\TQdeltac \in (0,1)$ is the ``gate'' parameter---i.e., the inflated probability of sampling a zero---and $\TQbc \in (0,1)$ is the success probability parameter under the standard Geometric distribution. We impose an ordering on both $C$-dimensional vectors of parameters $\TQdelta$ and $\TQb$ using the same transformed Ordered Normal prior as in the previous subsection:   

\begin{align}
    S^{-1}(\TQdelta) &\sim \ordnormal(\mu, \sigma) \\
    S^{-1}(\TQb) &\sim \ordnormal(\mu, \sigma) \label{eq:success_prob}
\end{align}
Importantly though, we reverse the ordering of $\vlambda$ such that for all $c \in \{1, \ldots ,C\}: \lambda_{c} < \lambda_{c-1}$. The reason is that higher $\TQdeltac$ and $\TQbc$ correspond, on average, to lower values sampled from the Zero-inflated Geometric and thus lower casualty counts and lower event intensity.

\begin{figure}[t]
     \centering
     \includegraphics[width=0.95\linewidth]{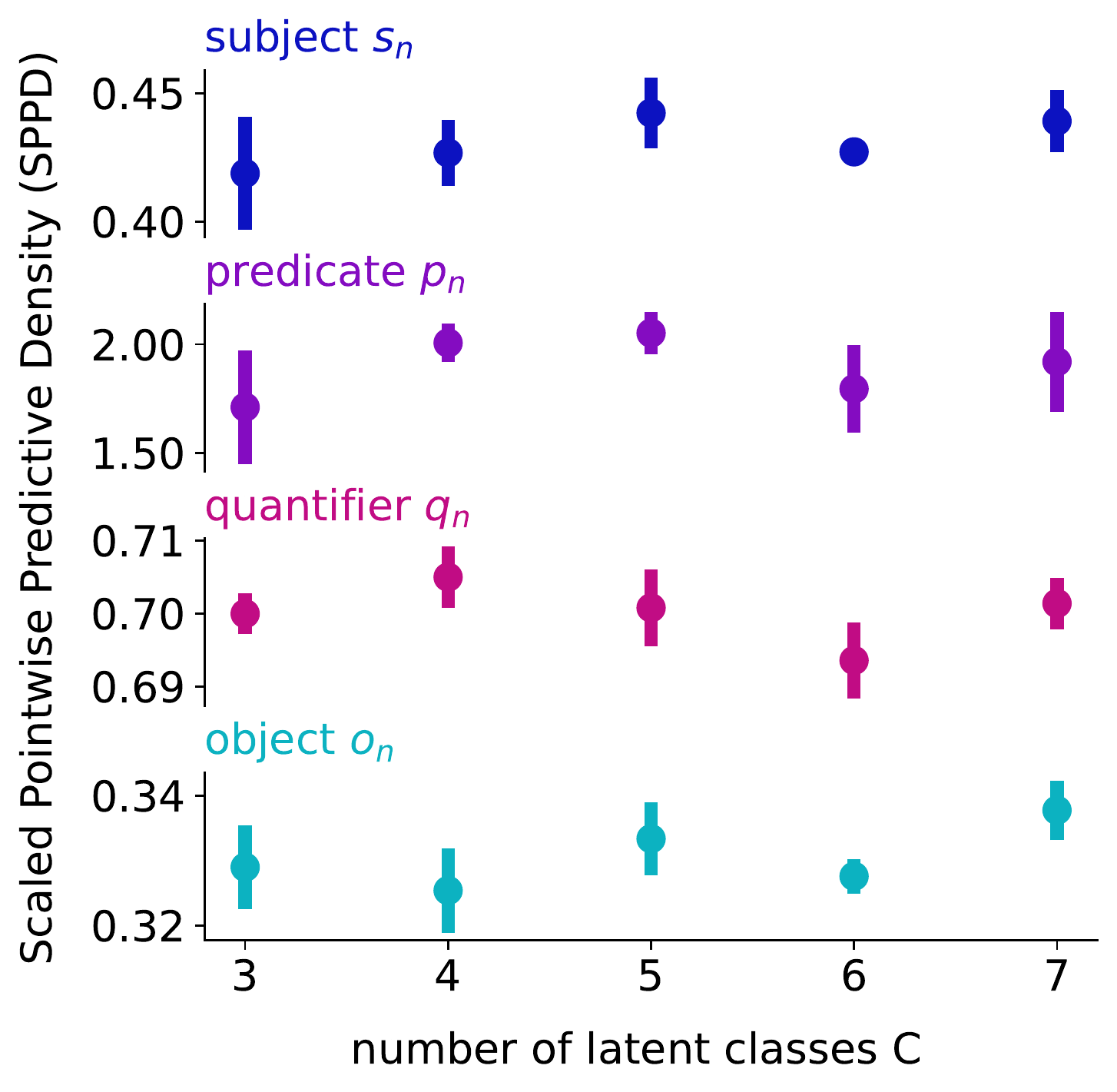} 
     \caption{Determining the optimal number of latent classes $\bC$. The values show the mean and standard deviation of the scaled pointwise predictive density (\sppd) on the held-out dataset over \num{5} random seeds. A model with $\bC = 5$ latent classes fits the data best overall.
     }
     \label{fig:optimal_c}
 \end{figure}

\subsection{Posterior Inference}
\label{sec:technical_details}

We implement the model using the probabilistic programming framework Pyro~\citep{bingham_pyro_2018}. Pyro provides an \href{https://docs.pyro.ai/en/stable/_modules/pyro/distributions/transforms/ordered.html}{\texttt{OrderedTransform}} function that implements~\cref{eq:orderedTransform}. We approximate the posterior distribution of our model's parameters using the No-U-Turn Sampler \citep[NUTS;][]{homan_no-u-turn_2014}, a variant of Hamiltonian Monte Carlo. NUTS is gradient-based and requires the parameters to be continuous. However, we explicitly model our latent variable $\sZ$ to be ordinal. To solve this problem, Pyro provides \href{https://pyro.ai/examples/enumeration.html#Mechanics-of-enumeration}{\texttt{parallel\_enumeration}} which marginalizes out discrete variables numerically during inference. We refer to all continuous parameters of our model as $\theta = \{\TZ, \TS, \TPomega, \TPkappa, \TQdelta, \TQb, \TO \}$ and to the full dataset of event tuples as $\bX$. NUTS produces samples of $\theta$ from the posterior $\theta^{(t)} \sim p(\theta \mid \bX)$. We can further sample the ordinal latent $\sZ^{(t)} \sim p(\sZ \mid  \theta^{(t)}, \bX)$ using Pyro's \href{https://docs.pyro.ai/en/dev/inference_algos.html#pyro.infer.discrete.infer_discrete}{\texttt{infer\_discrete}}.  Based on the samples $\{\sZ^{(t)}\}_{t=1}^T$ per event $\sN$, we can then compute a point estimate of the event's ordinal intensity either by taking the mean $\sZbar = \frac{1}{T} \sum_{t = 1}^{T} \sZ^{(t)}$ or the mode (i.e., the most frequently sampled class) $\sZdot$.

\paragraph{Label Switching.} It is typically not meaningful to compute $\sZbar$ for mixture and admixture models due to the problem of label switching~\citep{stephens_dealing_2000}, where the labels of the latent classes may switch between Markov chain Monte Carlo (MCMC) iterations. However, in the proposed model, the ordering transformation presented in \cref{eq:orderedTransform} represents an identifiability constraint that fixes the meaning of $\sZ=c$---i.e., $c$ is the class that is more intense than $c-1$ and less intense than $c+1$. This alleviates the problem of label switching, and permits us to meaningfully average posterior samples $\sZ^{(t)}$.

\paragraph{Practical Details.} Our model has $13 \times \bC$ trainable parameters, where $\bC$ is the number of latent classes. When fit to the full dataset introduced in \cref{sec:data}, our  implementation generates 1,000 posterior samples, for $C=5$, on a CPU with \num{64} GB of RAM in less than \num{10} minutes. Throughout we set hyperparameters to uninformative values: $\mu = -1.0$,  $\sigma = 1.0$, $k = 1.0$ and $\eta = 1.0$.

\begin{figure*}[t]
 \centering
 \includegraphics[width=1.0\linewidth]{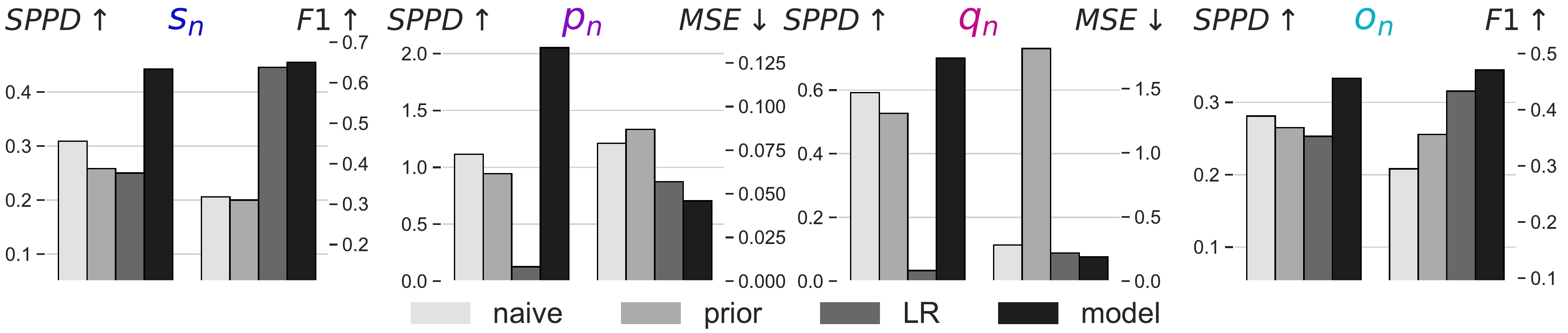} 
\caption{Intrinsic evaluation on the held-out set: we impute removed values $\sY$, e.g., $\{\sP\}$, of one site based on values $\sYremain$ of remaining sites, e.g., $\{\sS, \sQ, \sO\}$. We present mean results of scaled pointwise predictive density (\emph{\sppd}) and error metrics (\emph{weighted F1} / \emph{MSE}) over \num{5} runs. We compare three variants of our model with $\bC = 5$ latent classes: a \emph{naive} baseline fitted only to the single site that is removed at test time; an unfitted version, whose parameters are randomly drawn from the \emph{prior}, and a fitted variant of the proposed \emph{model}. In addition, we train four linear regression (\emph{LR}) models to directly predict $\sY$ from $\sYremain$.}
\label{fig:results}
\end{figure*}
 
\section{Intrinsic Evaluation: Imputation}
\label{sec:intrinsic_evaluation}

To evaluate the fit of our model, we conduct a series of predictive experiments. We randomly split the $N=\num{112089}$ events into a \num{70}\% training and \num{30}\% held-out set. We fit the model on the training set $\bX$ and approximate the posterior distribution of continuous parameters $\theta$ based on $T = \num{1000}$ NUTS samples after discarding the first \num{200} burn-in samples. For evaluation on the held-out set $\bY$, we report scaled pointwise predictive density (\sppd), mean squared error and weighted F1 score as detailed in \cref{sec:metrics}.

\subsection{Number of Latent Classes $\bC$}
\label{sec:optimal_c}

The number of classes $\bC$ that our latent variable $\sZ$ ranges over can be regarded a hyperparameter. To identify the optimal setting of $\bC$, we repeatedly fit the model on the training set and evaluate it on the held-out set. We do so for each class setting in $\bC = [3, \ldots, 7]$ and find that $\bC = 5$ yields the best model fit overall, as shown in \cref{fig:optimal_c}.

\subsection{Imputation Task}
\label{sec:imputation}

\paragraph{Procedure.} 

Next, we conduct a data imputation task. On the held-out set, we remove one observed site from all event tuples and refer to it as $\sY$ --- e.g., we remove all Goldstein values and obtain $\sY = \{ \sP \}$. We use the remaining three-way event tuples $\sYremain = \{ \sS, \sQ, \sO \}$ jointly with the posterior of $\theta$ to make predictions of $\sY$. In particular, we draw samples from the posterior predictive distribution $\sYhat^{(t)} \sim p(\sY \mid \theta^{(t)}, \sYremain)$ and compute their mean $\sYhat = \frac{1}{T} \sum_{t = 1}^{T} \sYhat^{(t)}$ to get point estimates of $\sY$.

\paragraph{Baselines.} We compare the proposed model against three baselines, where the first two baselines are simple and we expect our proposed model to outperform them. For the first baseline, we fit four versions of the proposed model to only one of the observed sites $\{ \sS, \sP, \sQ, \sO \}$. This naive baseline effectively just learns the empirical histogram of each site in the training set, and uses the relevant one to make predictions for $y_n$ --- we refer to this as \emph{naive}. For the second baseline, we use an unfitted version of the proposed model, whose parameters have been sampled from the \emph{prior}. For the last baseline, we train four linear regression models (\emph{LR}) to predict $\sY$ directly from $\sYremain$.

\paragraph{Results.} 
\cref{fig:results} reports the results of the imputation experiments, where the proposed model performs comparatively well, obtaining substantially better predictive performance over both the naive and non-naive baselines.

\begin{table*}[t]
\fontsize{11}{11}\selectfont
\centering
\renewcommand{\arraystretch}{1.4} 
\setlength{\tabcolsep}{0.50em} 
\begin{tabular}{cccccc?ccccc}
\multicolumn{1}{l}{}                & \multicolumn{5}{c?}{\textbf{forecasting Goldstein time series $\bPvec$}}                                                                                                        & \multicolumn{5}{c}{\textbf{forecasting casualty count time series $\bQvec$}}                                                                                                     \\
\multicolumn{1}{c|}{\textbf{}}      & \multicolumn{1}{c|}{$\bPvec \xrightarrow{} \bPvec$} & \multicolumn{2}{c|}{$\bPvec, \bQvec \xrightarrow{} \bPvec$} & \multicolumn{2}{c?}{$\bPvec, \bZvec \xrightarrow{} \bPvec$} & \multicolumn{1}{c|}{$\bQvec \xrightarrow{} \bQvec$} & \multicolumn{2}{c|}{$\bPvec, \bQvec \xrightarrow{} \bQvec$} & \multicolumn{2}{c}{$\bQvec, \bZvec \xrightarrow{} \bQvec$} \\
\multicolumn{1}{c|}{country $\ell$} & \multicolumn{1}{c|}{MSE}                            & MSE                    & \multicolumn{1}{c|}{Grang}         & MSE                              & Grang                    & \multicolumn{1}{c|}{MSE}                            & MSE                    & \multicolumn{1}{c|}{Grang}         & MSE                              & Grang                   \\ \hline
\multicolumn{1}{c|}{Egypt}          & \multicolumn{1}{c|}{3.33}                           & 3.15                   & \multicolumn{1}{c|}{0.30}          & \textbf{3.14}                    & 0.19                     & \multicolumn{1}{c|}{0.22}                           & 0.15                   & \multicolumn{1}{c|}{0.79}          & \textbf{0.14}                    & 0.35                    \\
\multicolumn{1}{c|}{Iraq}           & \multicolumn{1}{c|}{4.85}                           & \textbf{4.82}          & \multicolumn{1}{c|}{0.05}          & \textbf{4.82}                    & 0.05                     & \multicolumn{1}{c|}{2.91}                           & \textbf{2.82}          & \multicolumn{1}{c|}{0.05}          & \textbf{2.82}                    & 0.14                    \\
\multicolumn{1}{c|}{Syria}          & \multicolumn{1}{c|}{\textbf{2.96}}                  & 3.10                   & \multicolumn{1}{c|}{0.29}          & 3.10                             & 0.35                     & \multicolumn{1}{c|}{0.56}                           & \textbf{0.49}          & \multicolumn{1}{c|}{0.28}          & 0.50                             & 0.10                    \\
\multicolumn{1}{c|}{Yemen}          & \multicolumn{1}{c|}{\textbf{5.09}}                  & 5.12                   & \multicolumn{1}{c|}{0.15}          & 5.12                             & 0.46                     & \multicolumn{1}{c|}{1.41}                           & \textbf{1.02}          & \multicolumn{1}{c|}{0.15}          & \textbf{1.02}                    & 0.02                   
\end{tabular}
\caption{Forecasting: we test how much predictive information our intensity time series $\bZvec$ holds on future developments of Goldstein values $\bPvec$ and casualty counts $\bQvec$. In most cases, the vector autoregression (VAR) outperforms the autoregression in terms of mean squared error (\emph{MSE}) (results reported are $10^{-1}$) on the held-out set. We find that $\bZvec$ holds information on both $\bPvec$ and $\bQvec$ as including $\bZvec$ as an additional time series performs en par with including $\bPvec$ or $\bQvec$ throughout the VAR experiments. If the $p$ value of the corresponding Granger (\emph{Grang}) test is small, we can reject the null hypothesis that the additional time series does \emph{not} add predictive information.}
\label{tab:forecasting}
\end{table*}

\section{Extrinsic Evaluation: Time Series}
\label{sec:extrinsic_evaluation}

In the last section, we intrinsically evaluated the proposed model by having it impute held-out data. In this section, we consider extrinsic evaluations of the proposed model, which seek to evaluate it on downstream predictive tasks for which it was not explicitly designed. In this section, we rely on features of events coded by NAVCO which our model does not access during training. In particular, NAVCO codes for the location $\ell_n$ and the timestamp $\tau_n$ of each event. We discretize time so that each event is associated with a specific month $m_n$. Augmented with these extra characteristics, as well as with the mean intensity inferred by our model, each event $n$ has the following attributes: $\{\sS, \sP, \sQ, \sO, \ell_n, m_n, \sZbar\}$. In all of the tasks in this section, we split events according to their location $\ell_n = \ell$ and examine monthly-aggregated time series of the cardinal and ordinal quantities $\{\sP, \sQ, \sZbar\}$. In particular, we ask two questions: 1) whether $\sZbar$ adds predictive information for forecasting $\sP$ and $\sQ$, and 2) whether the model-based measure of intensity $\sZbar$ correlates with Google Trends more than just $\sP$ and $\sQ$.\looseness=-1

\paragraph{Event Time Series.}
We construct time series in the following way. We first run inference on a training dataset $\bX$ to obtain posterior samples of the parameters $\{\theta^{(t)}\}_{t=1}^T$. We then use the trained parameters to obtain intensity scores $\{\boldsymbol{z}^{(t)}\}_{t=1}^T$ for both events in the training set as well as events in a test set that the model did not access. We then compute $\bar{z}_n$ for each $n$ and re-scale the $\bar{z}_n$ estimates to cover a $[0,1]$ range as we did with Goldstein values. For each location $\ell$ and month $m$ appearing in the data, we then construct the posterior average intensity class $\bar{z}_{\ell, m}$:\looseness=-1
\begin{equation}
\bar{z}_{\ell, m} = \frac{\sum_{n=1}^N  \bar{z}_n\, \mathbbm{1}(\ell_n = \ell) \mathbbm{1}(m_n = m)}{\sum_{n=1}^N \mathbbm{1}(\ell_n = \ell) \mathbbm{1}(m_n = m)}
\end{equation}
For a given location $\ell$, we refer to the full time series over months as $\bZvec$ and linearly interpolate the entries $m$ corresponding to months for which there is no data. We similarly aggregate the observed sites per event to obtain time series of predicates $\bPvec$ and quantifiers $\bQvec$.

\begin{figure*}[t]
     \centering
     \includegraphics[width=1.0\linewidth]{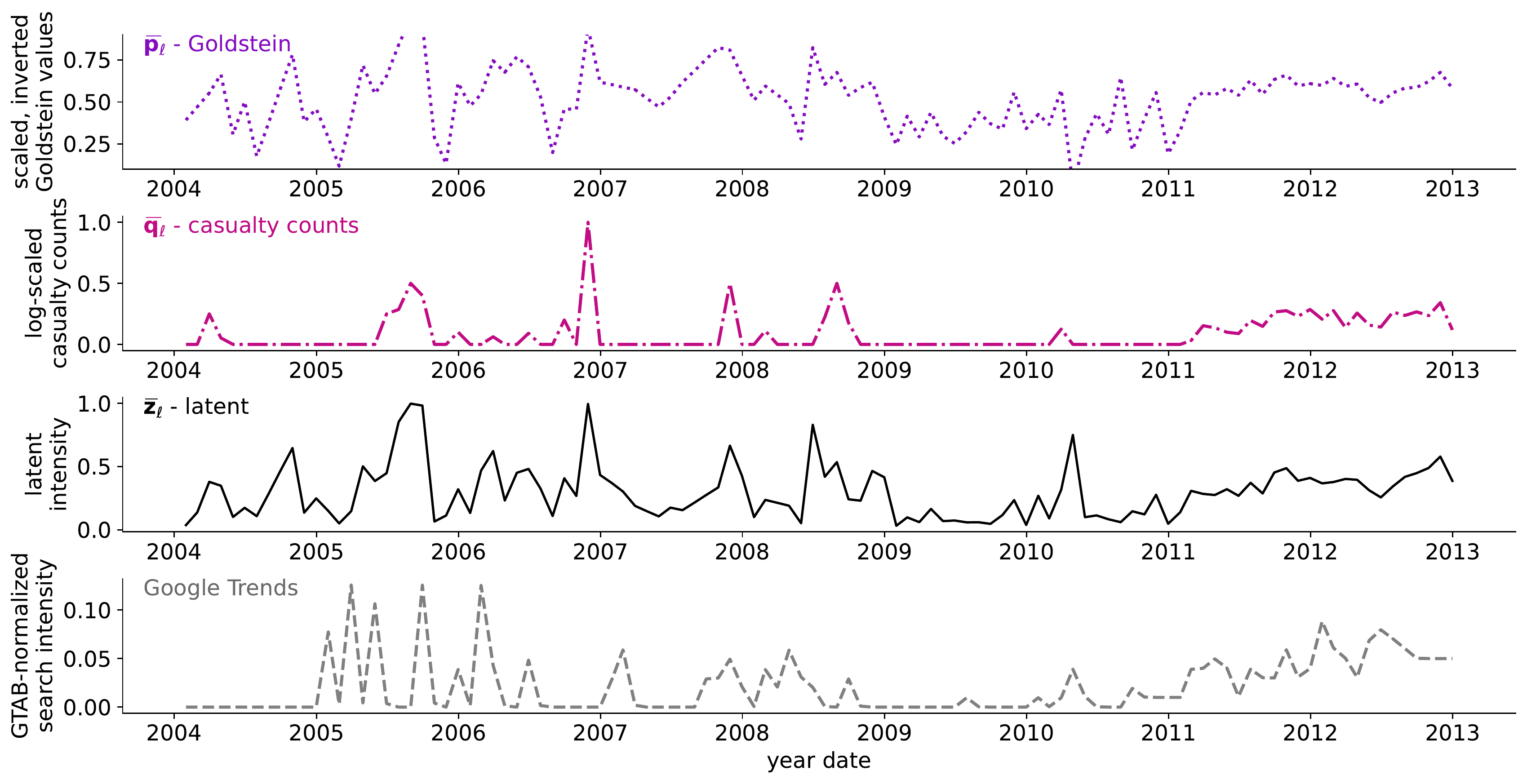} 
     \caption{Time series of events in Syria between 2004 and 2013, all values averaged by month. We observe that our intensity time series $\bZvec$ smoothly interpolates between the Goldstein $\bPvec$ and casualty counts time series $\bQvec$. We look at Google trends as an external correlate for (perceived) conflict intensity and report correlations in \cref{fig:heatmap_month}.}
     \label{fig:time_series_month}
 \end{figure*}

 \begin{figure}[h]
     \centering
     \includegraphics[width=1.0\linewidth]{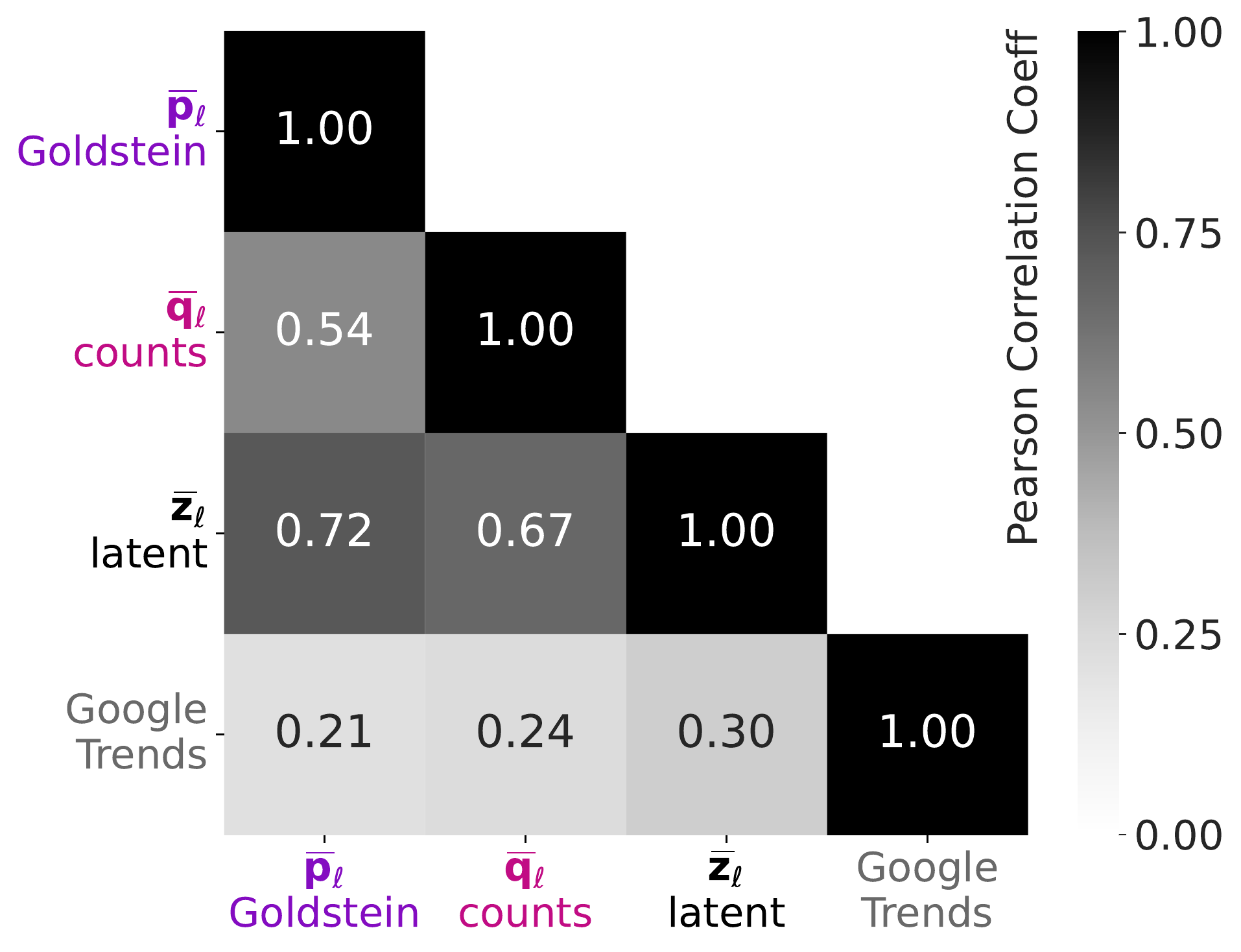} 
     \caption{Pearson correlation between time series displayed in \cref{fig:time_series_month}. Our latent intensity time series $\bZvec$ is positively correlated with the Goldstein $\bPvec$ and casualty counts time series $\bQvec$. Google trends are more strongly correlated with $\bZvec$ than with $\bPvec$ and $\bQvec$.}
     \label{fig:heatmap_month}
 \end{figure}
 
\paragraph{Autoregressive Forecasting.} Does knowledge of the inferred intensity time series $\bZvec$ improve forecasting of the Goldstein $\bPvec$ and casualty count $\bQvec$ time series? To test this, we first consider autoregressive (AR) models that forecast monthly values of each time series (e.g., $\Pdark{\bar{p}_{\ell, m}}$) based on the previous values of only that same time series (e.g., $\Pdark{\boldsymbol{\bar{p}}_{\ell, 1:(m-1)}}$). We then consider vector autoregressive (VAR) models that use multiple time series (e.g., $\Pdark{\boldsymbol{\bar{p}}_{\ell, 1:(m-1)}}$ and $\Qdark{\boldsymbol{\bar{q}}_{\ell, 1:(m-1)}}$) to form the same predictions. When incorporating values from our latent intensity time series $\bZvec$, we must be careful to avoid test set leakage. To do so, we fit our model to a subset of data $X_{\backslash \ell}$ that excludes any events in a particular location $\ell$. We obtain posterior samples $\{\theta^{(t)}_{\backslash \ell}\}_{t=1}^T$ by fitting the proposed model to $X_{\backslash \ell}$. We then use these parameters to obtain $\bar{z}_n$ for all events $n$ in both the training $X_{\backslash \ell}$ and test set $X_{\ell}$.

Before fitting any (vector) autoregressive models and performing the forecasting experiments, we apply a first-order differencing transform to all time series to remove potential linear trends, and verify each time series' stationarity using an Augmented Dickey-Fuller test \citep{fuller_introduction_1976}. We then fit an autoregressive (AR) model to each individual time-series and a vector autoregressive (VAR) model to all pairs of time series. We determine their optimal orders (lag) in months using Bayesian Information Criterion (BIC), and use cross validation to measure held-out forecasting error across 24 folds. In \cref{tab:forecasting}, we report results in mean squared error (MSE) on the held-out sets averaged over all \num{24} folds. Indeed, our results suggest that $\bZvec$ contains (predictive) information on both $\bPvec$ and $\bQvec$. For instance, in the VAR experiments for forecasting $\bPvec$, we may replace the additional time series $\bQvec$ with $\bZvec$ without any drop in performance. We also report Granger tests \citep{granger_investigating_1969} which test the null hypothesis that forecasting a variable (e.g., $\bPvec$) using only its own history is no less accurate than also using an additional variable's history (e.g., $\bZvec$).

\paragraph{Descriptive Analysis.}
A growing body of work analyzes the correlation between shifts in online behavior or media attention and conflict intensity \citep{chykina_using_2018, timoneda_spikes_2022}. Following this line of work, we download time series of Google search keywords using the \href{https://github.com/epfl-dlab/GoogleTrendsAnchorBank}{Google Trends Anchorbank (G-TAB)} \citep{west_calibration_2020}. We constrain searches to the category of ``World News'' and use the country name of location $\ell$ as the search keyword. In contrast to the forecasting setting, we fit our model to the entire dataset $\bX$, including events associated with the respective location, and obtain an intensity time series $\bZvec$. \cref{fig:time_series_month} shows a comparison between four time series: the Goldstein $\bPvec$, casualty counts $\bQvec$, our latent $\bZvec$ and the Google trends time series for Syria between 2004 and 2013.\footnote{\citet{fisher_google_2012} reports a correlation between Google search volume and activities in the Syrian Civil War for this period. 2004 is the earliest date to query Google trends.} We observe that $\bZvec$ nicely interpolates between the time series of $\bPvec$ and $\bQvec$. It captures larger trends and fluctuations in both time series, while not exactly mirroring either. Further, in \cref{fig:heatmap_month}, we report Pearson correlation coefficients between all pairs of time series. While $\bPvec$ and $\bQvec$ are positively correlated, $\bPvec$ is more strongly correlated with $\bZvec$ than with $\bQvec$. Moreover, Google trends are more strongly correlated with $\bZvec$ than either $\bPvec$ or $\bQvec$. We hypothesize this may be due to the additional information on the subject $\sS$ and object types $\sO$ that $\sZ$ encodes, and which may contribute to how much attention is paid to different conflicts, as measured by Google Trends.

\section{Related Work}
\label{sec:related_work}

There are a number of relevant papers that broadly seek to measure ``latent concepts''~\citep{douglass_introducing_2022} pertaining to international relations, particularly using event data. \citet{terechshenko_hot_2020} uses a Bayesian item-response theory model to learn ordinal conflict intensity levels from observed event types. \citet{oconnor_learning_2013} present an unsupervised, probabilistic topic model to learn an ontology of event data and use the Goldstein scale to evaluate it. \citet{schein_bayesian_2015, schein_bayesian_2016, schein_poisson-randomized_2019} decompose four-way tensors (senders, receivers, actions, time steps) to infer latent classes of CAMEO-coded events. \citet{stoehr_ordered_2023} build on those models by further imposing an ordering on their latent space, which captures conflict--cooperation intensity. Another line of work models friend--enemy relationship trajectories using neural network-based \citep{han_no_2019} or hidden Markov model-based \citep{chaturvedi_unsupervised_2017} approaches. There also exists work on signed network representations of relationships that are extracted from text \citep{srivastava_inferring_2016, choi_document-level_2016} or Wikipedia conflict articles \citep{stoehr_classifying_2021}.~\looseness=-1

\section{Future Work}

\label{sec:discussion}

Assuming civilian casualties are \setquote{more intense} than military casualties, we could impose an additional ordering on object types $\sO$. We could also condition observed sites on each other---e.g., casualty counts $\sQ$ are conditionally dependent on $\sO$ or even $\sP$ under the model. We plan to further incorporate multiple latent variables to model multi-dimensional intensity concepts. Future models could also condition on location and include a temporal component to account for how surprisal may affect the perceived intensity of an event. We also note that the latent variable model presented in this work could also be extended to a more general framework for learning interpretable, ordinal scales from a set of mixed-type data that include cardinal or ordinal observations. We plan to explore generalizations of the proposed model and applications beyond of international relations.

\section*{Acknowledgments}

We would like to thank Giuseppe Russo and Benjamin Radford for helpful discussions. Moreover, we acknowledge the feedback received from the anonymous reviewers, the NLP groups at ETH Zurich and the DLAB at EPFL. NS acknowledges support from the Swiss Data Science Center (SDSC) fellowship. LTH acknowledges support from the Michael Athans fellowship fund.

\section*{Limitations}

 We discuss limitations of our modeling assumptions in \cref{sec:discussion}. They are based on prior work in political science such as the CAMEO ontology and the Goldstein scale. On this account, they may replicate or potentially introduce biases. Hyperparameter search, settings and implementation details are provided in \cref{sec:technical_details} and \cref{sec:intrinsic_evaluation}. All NAVCO event descriptions are limited to English language, but do not disclose individuals. 

\section*{Impact Statement}

We emphasize that our models are intended for research, analysis and monitoring purposes. They should not be blindly deployed for automatized decision-making processes. The notion of conflict intensity is intrinsically hard to quantify: it is strongly dependent on socio-cultural background and subjective experience.

\bibliography{references}
\bibliographystyle{acl_natbib}

\appendix
\newpage

\section{Evaluation Metrics}
\label{sec:metrics}

To evaluate the posterior predictive distribution $p(\sY \mid \theta^{(t)})$, we consider a scaled variant of the log pointwise predictive density \citep[\lppd;][]{gelman_posterior_1996, gelman_understanding_2014}, which we term \emph{\sppd}:
\begin{align}
    \exp\left({\frac{1}{N} \sum_{n = 1}^{N} \log \big( \frac{1}{T} \sum_{t=1}^{T} p(\sY \mid \theta^{(t)}) \big)}\right)
\end{align}
The term inside the log, $\frac{1}{T} \sum_{t=1}^{T} p(\sY \mid \theta^{(t)})$,  computes a Monte Carlo approximation to the posterior predictive density for a given $y_n$. The rest, $\exp\left({\frac{1}{N} \sum_{n = 1}^{N} \log( \cdot )}\right)$, then computes the geometric mean of the pointwise densities.\looseness-1

We also compute point estimates via the posterior predictive mean $\frac{1}{T} \sum_{t = 1}^{T} \expect[y_{n} \mid \theta^{(t)}]$ which allows comparing predicted and true values based on error metrics like weighted F1 score or mean-squared error (MSE).

\section{Inverse of the \OrdNorm} 
\label{app:inv-ordered-normal}

We define the \OrdNorm distribution with the help of an ordering transformation in \cref{eq:orderedTransform}. This transformation is a smooth bijection since its inverse is given by:
\begin{align}
x_c \leftarrow
\begin{cases}
    \lambda_{1} & \text{if } c = 1\\
    \log(\lambda_{c} - \lambda_{c - 1}) & \text{if } c > 1
\end{cases}
\end{align}
Note that for all $c$, $\lambda_{c} > \lambda_{c-1}$, so the $\log$ is well-defined.

\begin{table*}[h]
\fontsize{11}{11}\selectfont
\centering
\renewcommand{\arraystretch}{1.5} 
\setlength{\tabcolsep}{1.00em} 
\begin{tabular}{l|l|l}
\textbf{CAMEO actor code} & \textbf{meaning}                      & \textbf{group 4 mapping} \\ \hline
REB                       & rebels                                & military                 \\
MIL                       & military                              & military                 \\
GOV                       & government                            & government               \\
ETH                       & ethnic                                & civilian                 \\
REL                       & religious                             & civilian                 \\
COP                       & police forces                         & military                 \\
JUD                       & judiciary                             & political                \\
OPP                       & political opposition                  & political                \\
LLY                       & regime loyalists                      & government               \\
ACT                       & activists                             & political                \\
NON                       & non-aligned third party               & military                 \\
SPY                       & state intelligence                    & military                 \\
UAF                       & unidentified armed forces             & military                 \\
UNS                       & unidentified unarmed non-state actors & civilian                 \\
NGO                       & non-governmental organisation         & political                \\
BUS                       & business                              & civilian                 \\
CVL                       & civilian group                        & civilian                 \\
IND                       & civilian individual                   & civilian                 \\
EDU                       & educators                             & civilian                 \\
STU                       & students                              & civilian                 \\
YTH                       & youth                                 & civilian                 \\
ELI                       & elites                                & civilian                 \\
LAB                       & labour                                & civilian                 \\
LEG                       & legislature                           & political                \\
PTY                       & political party                       & political                \\
MED                       & media                                 & civilian                 \\
REF                       & refugees                              & civilian                 \\
IGO                       & inter-governmental                    & political                \\
NGM                       & non-governmental movement             & political                \\
MNC                       & multinational cooperation             & civilian                 \\
INT                       & international actors                  & political                \\
TOP                       & top officials                         & political                \\
MID                       & mid-lower level officials             & political                \\
HAR                       & hardliners                            & political                \\
MOD                       & moderates                             & political               
\end{tabular}

\caption{CAMEO actor types to group 4 mapping. The NAVCO columns \setquote{actor3}, \setquote{actor6} and \setquote{target3}, \setquote{target6} correspond to actor types defined by the \href{http://data.gdeltproject.org/documentation/CAMEO.Manual.1.1b3.pdf}{CAMEO actor codebook}. We consider \num{33} generic CAMEO actor types and map them into one of the \num{4} classes \setquote{civilian}, \setquote{military}, \setquote{governmental} and \setquote{political}.}
\label{tab:actor_mappings}
\end{table*}
\end{document}